\documentclass[11pt]{article}
\usepackage{paralist}

\usepackage{acl}
\usepackage{booktabs}
\usepackage{multirow}
\usepackage{times}
\usepackage{latexsym}
\usepackage{amsmath} 
\usepackage{amssymb} 
\usepackage{svg}
\usepackage[T1]{fontenc}
\usepackage{csquotes}

\usepackage[utf8]{inputenc}

\usepackage{microtype}

\usepackage{inconsolata}

\usepackage{graphicx}

\title{Towards Full-Text Scholarly Argument Mining using Entity-Anchored Claim Extraction}

\author{
    Bhuvanesh Verma$^{1}$, 
    Andy L\"uking$^{2}$, 
    Mounika Marreddy$^{1}$, 
    Leon Lukas Hammerla$^{1}$, 
    Alexander Mehler$^{1}$ 
    \\
    $^{1}$Text Technology Lab (TTLab), Goethe University Frankfurt, Germany \\
    \texttt{\{verma,M.Marreddy,hammerla,mehler\}@em.uni-frankfurt.de} \\
    $^{2}$Fakultät für Germanistik, Technische Universität Chemnitz, Germany \\
    \texttt{andy.luecking@phil.tu-chemnitz.de}
}

\begin{document}
\maketitle
\begin{abstract}
Argument mining in full-text scholarly articles remains an important yet challenging task.
%
This difficulty stems not only from the need to model long-distance dependencies but also from the substantial effort required to annotate sentence-level argumentative components throughout lengthy documents.
%
%
%
To address these challenges, we introduce a novel approach that leverages scientific entity mentions (e.g., tasks, datasets, and models) to identify and structure claims in scholarly literature.
%
We validate our approach through a case study involving 50 full-text research papers on sentiment analysis.
%
We first establish a reliable human gold standard (Krippendorff's $\alpha = 0.80$), identifying 113 claim-bearing sentences and extracting 199 structured claim records from 300 candidate sentences.
%
To address the annotation bottleneck, we employ a majority-voting ensemble of large language models (LLMs) to generate a silver-standard dataset comprising 1,487 claims.
Encoders fine-tuned exclusively on the silver-standard dataset outperform those trained on a small human-annotated gold-standard dataset and demonstrate strong zero-shot transfer to NLP papers outside the training corpus.

\end{abstract}

\section{Introduction}
Scientific literature is a rich source of complex reasoning, making it a particularly relevant domain for scientific argument mining \cite{Teufel:et:al:1999,Al-Khatib:et:al:2021}.
%
The central objective of scientific argument mining (SAM) is to identify argumentative components and reconstruct their relations in scholarly texts \cite{Lawrence:Reed:2019,Al-Khatib:et:al:2021}.
%
Scholarly text analysis often draws on established argumentation frameworks, such as Toulmin's model \cite{toulmin1958uses} and Walton's argumentation schemes \cite{walton2008argumentation}, to identify claims, supporting evidence, and their argumentative relationships.
%
However, existing research has largely focused on abstracts or individual sections of scientific articles
\cite{accuosto2019transferring, mayer2020transformer, fergadis2021argumentation,accuosto2021argumentation}. 
%
One reason for this focus is the substantial effort required for full-text annotation, as illustrated by the Sci-Arg corpus \cite{lauscher2018argument}.
Furthermore, identifying argumentative components and their relationships across entire documents poses substantial challenges for neural models \cite{binder2022full}.

To address these two challenges, we draw on approaches from general-domain argument mining, particularly those applied to political discourse and social media, where arguments are often identified in relation to specific controversial topics \cite{levy2018towards,ein2020corpus,verma2026predicting}.
%
In these domains, topics serve as semantic anchors that facilitate sentence-level annotation independently of broader document contexts and enable systematic comparisons of claims across different sources.
%
Although scientific discourse exhibits comparable structures, previous research on SAM has paid relatively little attention to this anchoring strategy.
%
Building on recent advances in scientific entity extraction \cite{jain2020scirex,otto2023gsap}, we propose a novel approach to claim extraction from scholarly articles based on scientific entity anchoring.
%
In this approach, scientific entities (e.g., tasks, methods, and datasets) serve as structural anchors, playing a role analogous to that of \enquote{topics} in general-domain argument mining.
%
Anchoring arguments to scientific entities reformulates full-text SAM in two fundamental ways.
%
First, it reduces the scope of annotation to individual sentences, enabling claims to be identified and structured without requiring prior resolution of long-range discourse dependencies.
%
Second, it facilitates cross-document comparability by providing a common structural reference for claims concerning the same method and dataset, regardless of differences in their surface formulations.

\begin{figure}[t]
    \centering
    \includegraphics[width=\columnwidth]{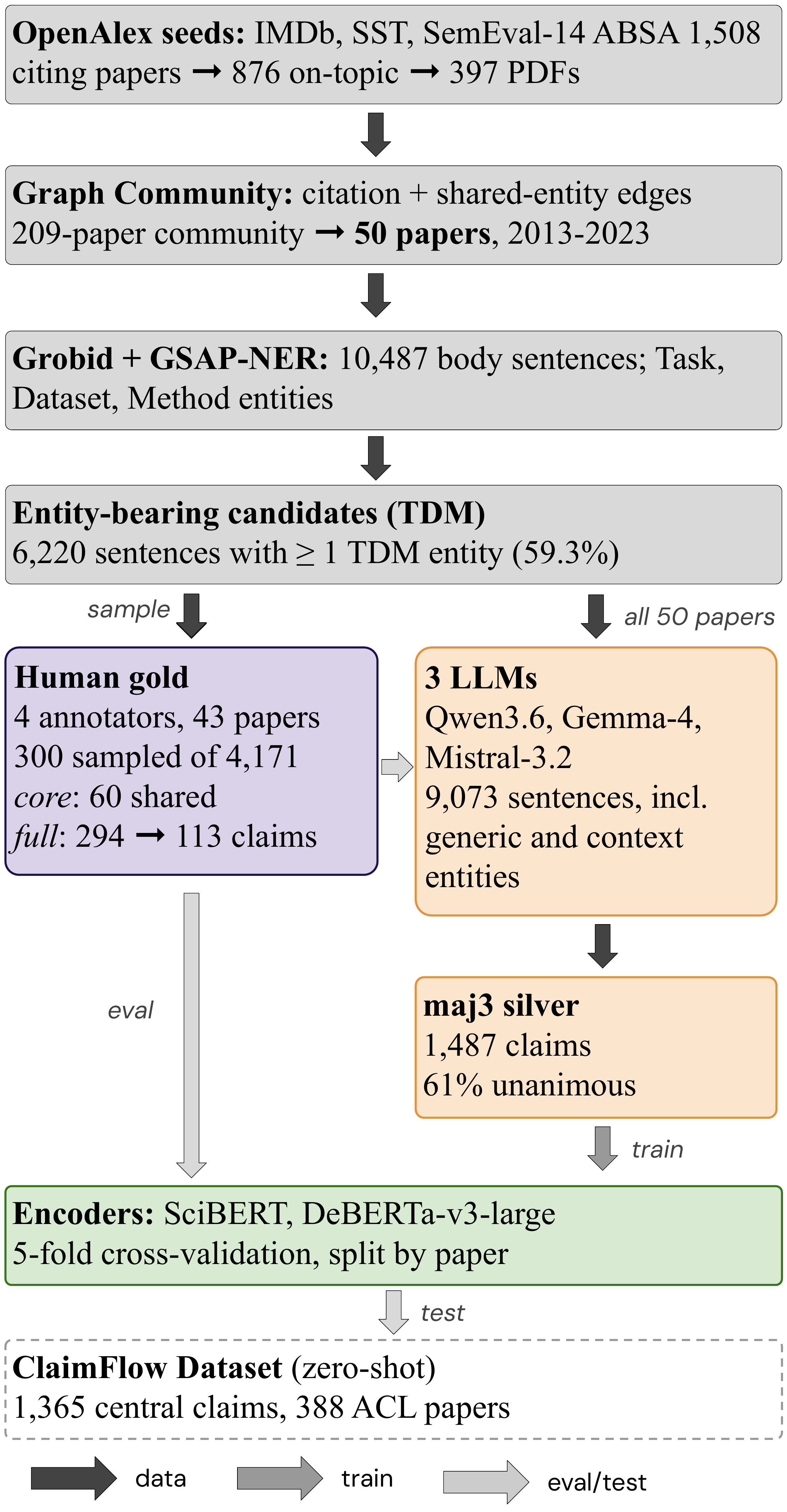}
    \caption{Pipeline of entity-anchored claim extraction.}
    \label{fig:sci-eam-pipeline}
\end{figure}

We present a pipeline for entity-anchored claim extraction from full-text scientific publications.
%
Rather than exhaustively annotating entire papers, we combine a carefully curated, independently human-annotated gold-standard dataset with a large-scale silver-standard dataset generated through consensus among three large language models (LLMs).
%
In a case study involving 50 research papers on sentiment analysis, we annotated 294 valid candidate sentences, identifying 113 claim-bearing sentences to establish a gold-standard dataset, which we subsequently used to validate a silver-standard dataset comprising 1,487 claims.
%
We demonstrate that encoders fine-tuned exclusively on the silver-standard dataset yield more stable claim identification than those trained on the smaller human-annotated gold-standard dataset or on data generated by a single LLM.
%
Furthermore, encoders trained on the silver-standard dataset demonstrate strong zero-shot generalization to NLP papers outside the case study corpus.
%
\autoref{fig:sci-eam-pipeline} provides an overview of the complete pipeline.

\section{Related Work}

\subsection{Full-Text Scientific Argument Mining}

\citet{teufel1999argumentative} introduced Argumentative Zoning (AZ) to classify sentences in scholarly articles by rhetorical function, such as background, aim, and contrast.
%
\citet{teufel-moens-2002-articles} implemented AZ using hand-crafted features and Naive Bayes classifiers, while subsequent extensions captured domain-specific rhetorical functions in chemistry (AZ-II) \citep{teufel2009towards} and biology \citep{mizuta2006zone}.
%
Similarly, \citet{liakata2012automatic} introduced the Core Scientific Concepts (CoreSC) scheme to classify sentences in full-text biochemistry articles into 11 categories using SVMs and CRFs, while \citet{fisas2016multi} developed the Dr.\ Inventor corpus to annotate discourse structure and novelty in computer graphics papers.
%
While rhetorical zoning captures the discourse structure of scholarly articles, identifying fine-grained argumentative components, particularly claims and supporting evidence, requires more specialized annotation schemes.
%
\citet{blake2010beyond} introduced a rule-based framework for distinguishing explicit, implicit, and comparative claims.
%
Building on the Dr.\ Inventor corpus, \citet{lauscher2018argument} introduced Sci-Arg, extending its annotations with argumentative components (\textit{Own Claim}, \textit{Background Claim}, \textit{Data}).
Following this line of work, we adopt \textit{Own Claim} and \textit{Reported Claim} as the primary argumentative components in our schema.

\subsection{Scientific Argument Mining using LLMs}

\citet{binder2022full} proposed a BERT-based pipeline for full-text argument mining, highlighting challenges in identifying non-contiguous argumentative units and their relations.
%
Although LLMs offer extended context windows that may help address these limitations, their application to scientific argument mining has primarily focused on claim verification and evidence-based reasoning.
%
Specifically, benchmarks such as SciFact-Open \citep{wadden2022scifact} and SciTab \citep{lu2023scitab} evaluate scientific claim verification against large literature collections and evidence from scientific tables, respectively.
%
Related approaches extend to peer-review analysis, assessing the evidential support for claims in reviews \citep{guo2023automatic} and linking reviewer feedback to subsequent manuscript revisions \citep{d2024aries}.
%
More recently, benchmarks such as CLAIM-BENCH \citep{javaji2025can} and CLAIMCHECK \citep{ou2025claimcheck} have evaluated LLMs on scientific claim-evidence reasoning and the grounding of peer-review critiques in scholarly articles, respectively.
%
Thus, while LLMs are increasingly used to verify and critique previously identified scientific claims, their potential for entity-anchored claim extraction from full-text scholarly articles remains comparatively underexplored.

\subsection{Entity-Anchored Scientific Information Extraction}
%
Scientific entity extraction has evolved from broad terminology extraction \cite{luan2019information} to fine-grained annotation schemes covering tasks, datasets, and metrics (TDMSci) \cite{hou2021tdmsci}, machine learning models and architectures (GSAP-NER) \cite{otto2023gsap}, and subsequently to the extraction of relations between scientific entities (SciER) \cite{zhang2024scier}.
%
In parallel, argument mining has increasingly incorporated topic-based approaches that anchor arguments to specific subjects of debate \cite{irani2024wiba}.
%
Building on these advances, MetaLead \cite{timmer2026metalead} introduces a richer annotation scheme linking tasks, training and test datasets, metrics, scores, and experiment types to support transparent comparisons of experimental results across scientific publications.
%
%
Structured scientific knowledge increasingly supports advanced tasks such as research idea generation. For example, SciMON \cite{wang2024scimon} leverages scientific concepts and prior literature to generate and iteratively refine novel research ideas.
%
While these extraction strategies have primarily served structured knowledge organisation or idea generation, their integration into full-text SAM remains limited.
We address this gap by using fine-grained scientific entities as explicit anchors for claim extraction.

%
%
%
%

\begin{figure}[t]
    \centering
    \includegraphics[width=\columnwidth]{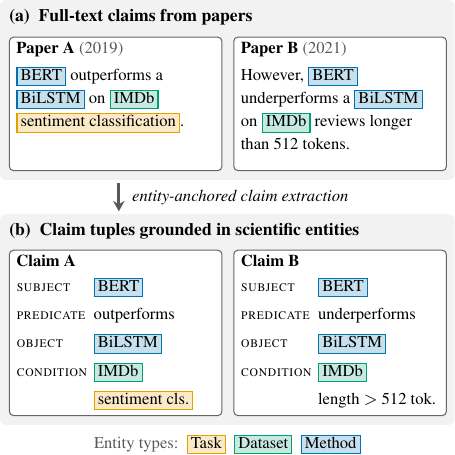}
    \caption{Extraction of claim tuples from full text, anchored to task, dataset, and method entities.}
    \label{fig:sci-eam-overview}
\end{figure}

\section{Entity-Anchored Claim Extraction}

We introduce \textbf{Entity-Anchored Claim Extraction} as the task of extracting claim structures from scholarly articles by anchoring them to scientific entities. 
%
Traditional approaches to SAM often treat every sentence or phrase as a potential claim \cite{teufel1999argumentative, lauscher2018argument}. 
%
In contrast, our approach focuses exclusively on sentences containing explicit mentions of scientific entities.
%
We define an \textit{entity-anchored claim} as a contestable and verifiable assertion about one or more scientific entities (e.g., tasks, datasets, metrics, or methods).
%
A sentence qualifies as claim-bearing if it makes an assertion about an entity's performance, capabilities, limitations, comparative advantages, or intrinsic properties, among others.

As illustrated in \autoref{fig:sci-eam-overview}, we represent each claim as a relational tuple in which scientific entities assume explicit roles (e.g., subject or object) linked by a predefined predicate.
%
Specifically, our controlled predicate inventory comprises four categories: 
performance comparisons (\texttt{outperforms}, \texttt{underperforms}, \texttt{is-comparable-to}), performance changes (\texttt{improves}, \texttt{degrades}, \texttt{achieves}), applicability and transfer (\texttt{generalizes-to}, \texttt{fails-on}, \texttt{is-suited-for}, \texttt{is-standard-for}), and dependencies and efficiency (\texttt{requires}, \texttt{is-more-efficient-than}). 
An additional \texttt{OTHER} label covers assertions outside this taxonomy.
%
We provide detailed predicate definitions and argument constraints in \autoref{tab:predicates} in the appendix.
%
%
Furthermore, following established argument mining schemes \cite{teufel1999argumentative,lauscher2018argument}, we distinguish \textbf{Own Claims}, concerning the authors' work, from \textbf{Reported Claims}, attributed to prior research.


\section{Case Study: Sentiment Analysis}
This case study evaluates the effectiveness of our entity-anchored approach to claim extraction from scholarly articles.
%
We investigate whether this approach enables standardized claim representations that lays foundation for the identification of implicit argumentative relations across scholarly articles beyond those captured by citation-based analysis.
%
This evaluation requires a corpus of densely interconnected scientific publications.
%
We therefore focus on sentiment analysis, a well-established research area in NLP with an extensive citation network.

\subsection{Corpus Construction}
To construct a densely interconnected corpus of sentiment analysis papers, we seeded our search with three foundational benchmark publications: IMDb \cite{maas2011learning}, SST \cite{socher2013recursive}, and SemEval-14 ABSA \cite{pontiki-etal-2014-semeval}.
%
Using OpenAlex \cite{priem2022openalex}, we retrieved papers citing these benchmarks and filtered them by domain-specific keywords and open-access availability.
%
To ensure dense conceptual connectivity, we represented the retrieved papers as a graph, connecting papers through direct citations or shared scientific entities drawn from a curated list of datasets and methods.
%
Using community detection \cite{clauset2004finding}, we identified a densely connected cluster of 209 papers.
%
From this community, we selected 50 papers using stratified sampling by publication year to ensure temporal diversity.
%
We parsed the documents using GROBID \cite{grobid}.
%
Our preprocessing pipeline preserves all in-text citations (2,274 author-year spans), which serve as essential anchors for identifying \textbf{Reported Claims}.
%
We provide further details on document retrieval, community detection, and text normalization in \autoref{app:corpus_stats}.

\subsection{Scientific Entity Extraction}
To anchor claims within our corpus, we first require robust extraction of scientific entity mentions. 
We utilize GSAP-NER \cite{otto2023gsap}, which provides a fine-grained taxonomy specifically tailored to machine learning models, datasets, and tasks. 
Following the original architecture, we train two SciDeBERTa-CS taggers \cite{jeong2022scideberta}, one for specific entity types and one for generic references, on the complete set of 100 GSAP-NER publications. 

For our 50-paper corpus, we retain entity mentions from six categories: \textit{Task}, \textit{Dataset}, \textit{Data\-sour\-ce}, \textit{Method}, \textit{MLModel}, and \textit{ModelArchitecture}.
%
We exclude generic mentions and URLs but retain citation markers (\textit{ReferenceLink}) 
to link reported claims to cited publications.
%
Our analysis shows that 59.3\% of body-text sentences contain at least one scientific entity, providing a substantial candidate pool for entity-anchored claim extraction.
%
We provide extraction metrics, post-pro\-ces\-sing rules, and corpus-level entity statistics in \autoref{app:sci_entity}.

\subsection{Annotation Details}
Four expert annotators (two doctoral candidates and two postdoctoral researchers), all specializing in computational linguistics, conducted the annotation.
%
We sampled a subset of 300 entity-anchored sentences for the claim detection task.
%
To balance workload distribution and enable reliable inter-annotator agreement (IAA) estimation, we assigned 120 instances to each annotator.
%
All four annotators independently annotated a shared set of 60 instances, which we used to compute IAA.
%
To ensure reproducibility, we sampled instances using a fixed random seed (\texttt{20260819}) 
%
and developed a dedicated annotation tool (\autoref{app:appendix_annotation_ui}) to support the process.

\paragraph{Claim Identification and Structuring.}
Before the main annotation phase, we conducted a pilot study involving seven papers to familiarize annotators with the annotation scheme and tool.
%
From the remaining 43 papers, we identified 4,171 eligible candidate sentences, each containing at least one scientific entity mention in the body text.
%
We stratified the sampling by document structure and selected 300 sentences.
%
Specifically, we oversampled sentences from the abstract, conclusion, and results sections.
%
Conversely, we undersampled sentences from the introduction, related work, and methodology sections, which often contain procedural descriptions, standard definitions, and generic contribution statements that fall outside our claim definition.

The annotation process proceeds in two stages. 
%
First, annotators determine whether the target sentence contains a claim.
%
If the sentence contains a claim, annotators classify it as either \textit{own\_claim} or \textit{reported\_claim}.
%
Second, annotators structure each claim by identifying its subject and object and selecting a predicate from a predefined inventory of 13 predicates (see \autoref{app:appendix_annotation_ui}). 
%
For reported claims, annotators additionally link each claim to its corresponding citation marker.

\section{Experiments}
Although entity-anchored claim extraction reduces annotation effort, it still requires considerable time and domain expertise.
%
To reduce this annotation burden, we investigate automated entity-anchored claim extraction using zero-shot LLMs and fine-tuned pretrained encoders.

\subsection{LLM-Based Extraction Pipeline}
We benchmark three LLMs: Qwen3.6-27B, Gemma-4-31B-it, and Mistral-Small-3.2-24B.
%
To ensure comparability, we use a common prompt template across all models, iteratively refined using Qwen3.6-27B.
%
We extract claims in three passes:

\begin{enumerate}
    \item \textit{Document Profiling (Paper Card):} 
    The LLM processes the title, abstract, introduction, section headings, and conclusion to generate a \textit{paper card} summarizing the paper's major claims, proposed methods, tasks, datasets, and core contributions.
    %
    %
    \item \textit{Sentence-Level Extraction:} The model processes target sentences in the context of their full sections to facilitate anaphora resolution.
    %
    The LLM determines whether a sentence contains a claim and, if so, extracts its subject, object, and predicate according to our manual annotation scheme.
    \item \textit{Rule-Based Coreference:} 
    A deterministic post-processing step replaces generic self-references (e.g., \enquote{we}, \enquote{our model}) with the explicit method name extracted from the paper card.
\end{enumerate}

%
%

\subsection{Fine Tuning via Silver Data}
While our zero-shot LLM pipeline identifies claims, we further investigate whether smaller, specialized encoders can achieve comparable performance more efficiently.
%
To this end, we aggregated sentence-level claim predictions from all three LLMs across the 50-paper corpus.
%
Using majority voting, we constructed a silver-standard dataset for claim identification from the three LLMs' predictions.
%
We then fine-tuned two pretrained masked language models, SciBERT \cite{beltagy2019scibert} and DeBERTa-v3-large \cite{he2021debertav3,he2021deberta}, with sequence classification heads using this dataset.
%
We employ five-fold cross-validation at the paper level, fine-tuning a separate model in each fold on silver-standard sentences from the training papers and evaluating it on gold-standard sentences from held-out papers.

\subsection{External Validation on ClaimFlow}
To determine whether our entity-anchored claim extraction strategy generalizes beyond our initial case study, we sought an external dataset for validation. 
%
Existing full-text argument mining corpora are either relatively dated \cite{teufel1999argumentative,teufel2009towards} or focus on domains outside NLP \cite{mizuta2006zone,blake2010beyond,liakata2012automatic,lauscher2018argument}, limiting their suitability for evaluating our scientific entity extraction model.
%
Recent resources also provide full-text claim annotations, including CLAIM-BENCH \cite{javaji2025can}, which links scientific claims to supporting evidence, and ClaimFlow \cite{pramanick2026claimflow}, which traces claims and their relationships across NLP publications.
%
We select ClaimFlow for its scale, NLP domain coverage, and sentence-level annotations, which align with our annotation granularity.
%
ClaimFlow identifies central claims in NLP papers, focusing on their abstracts, introductions, and conclusions.
%
The dataset provides paraphrased claims alongside supporting evidence sentences from the source text.
%
We use ClaimFlow claims to evaluate our fine-tuned encoder models.

\section{Results and Discussion}
We present the results of our annotation study and evaluate the performance of LLMs and fine-tuned encoders.
%
%
Unless otherwise specified, we report 95\% confidence intervals (CIs) for all annotation and model evaluation results, using 10,000 bootstrap resamples.

\subsection{Annotation Analysis}

We evaluate inter-annotator agreement for claim detection on 60 \textit{core} sentences.
Excluding the annotator who worked on annotation guidelines, the three independent annotators achieved a reliable Krippendorff's $\alpha$ of 0.80 (95\% CI: 0.66-0.92).
%
%
Across the full dataset of 300 sentences we use annotation from all four annotators.
Out of 300, 6 sentences were discarded as unusable, 113 (38.4\%) were identified as claims, predominantly concentrated in the results sections and beyond. 
%
For the claim structure extraction, annotators produced 199 records across 113 distinct \textit{gold} claim sentences, heavily anchored on scientific entities for both subjects (90.5\%) and objects (66.3\%).
%
%
Annotators agreed on the number of claims per sentence in 93\% of shared cases. However, agreement was moderate for subjects ($\alpha = 0.64$, 95\% CI: 0.46--0.78) and objects ($\alpha = 0.65$, 95\% CI: 0.45--0.81), but low for predicates ($\alpha = 0.46$, 95\% CI: 0.23--0.66).

Disagreement in subject and object annotations primarily arose from different interpretations of complex, multi-claim sentences. 
%
Additionally, subject mismatches frequently resulted from differing preferences in entity span granularity (e.g., selecting simply \textit{PBAN} versus the more expansive \textit{PBAN + fine-grained + bidirectional attention}). 
%
Conversely, the majority of predicate disagreements (56\%) were driven by the \texttt{OTHER} class. 
For example, given the sentence \enquote{syntactic target information derived from parses complements well with the left-target-right context representation}, one annotator selected \texttt{OTHER} (specifying \textit{complements}), while the remaining two attempted to utilize the predefined taxonomy, selecting \texttt{is-suited-for} and \texttt{is-comparable-to}, respectively.
%
Overall, \textit{outperforms} (24\%) and \textit{achieves} (22\%) were the most frequent predicates, while negative or limiting predicates accounted for only 11\%.
%
Additionally, 95\% of claim records were classified as \textit{own\_claim}.
\begin{figure}[t]
    \centering
    \includegraphics[width=\columnwidth]{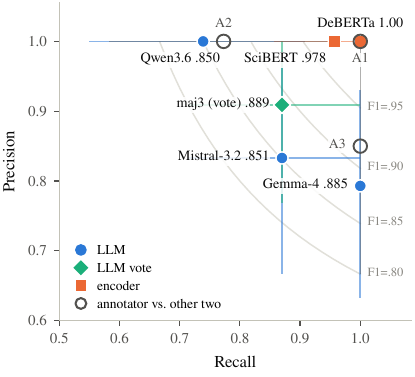}
    \caption{Performance on the 60 \textit{core} sentences evaluated against the human majority label. Whiskers indicate 95\% bootstrap CIs, and grey curves denote constant F1 contours. Open circles (A1--A3) show individual annotator agreement against the majority of the other two.}
    \label{fig:core_eval}
\end{figure}

\subsection{Model Evaluation}

We evaluated the three LLMs and their majority-vote ensemble, denoted as \textit{maj3}, on both the 60 shared sentences (\autoref{fig:core_eval}) and the full 294-sentence gold-standard set (\autoref{tab:llm}).
%
On the \textit{core} subset, \textit{maj3} achieved the highest F1 score (0.889) among the LLMs and showed strong agreement with the majority human annotation ($\kappa = 0.82$), comparable to that of the least-consistent annotator with the other two ($\kappa = 0.81$).
%
On the full gold-standard set, \textit{maj3} achieved the highest agreement ($\kappa = 0.78$) and F1 score (0.864) among the LLMs, although its F1 score was statistically indistinguishable from Gemma's and significantly higher only than Qwen's.
%
Overall, all models tended to identify more claims per sentence than human annotators, resulting in lower precision.
%
Despite moderate human agreement on claim structure, we evaluated the generated structures against gold-standard annotations for all correctly identified claim-bearing sentences.
%
Qwen achieved the highest F1 scores for all three structural components: subjects (0.79), predicates (0.60), and objects (0.55).
%
Mistral followed with F1 scores of 0.74, 0.53, and 0.47, respectively, while Gemma performed worst (0.71, 0.51, and 0.43).
%

Given the strong performance of \textit{maj3}, we applied it to all 9,073 sentences in our 50-paper corpus, yielding a silver-standard dataset comprising 1,487 claims (\autoref{tab:dataset_statistics}). 
%
These sentences include all annotated entities (generic or specific) and indicate whether a specific entity occurs in either of the two preceding sentences.
%
We train BERT-based encoders on this silver-standard dataset and evaluate them against the smaller human-annotated gold-standard dataset (\autoref{tab:llm}).
%
Averaged over three random seeds, DeBERTa-v3-large achieves an F1 score of 0.877 ($\pm$ 0.008), slightly exceeding SciBERT (0.865 $\pm$ 0.014), although the difference is not statistically significant.
%
Its performance also does not differ significantly from that of the individual LLMs or the teacher ensemble, \textit{maj3}.
%
Despite the small gold-standard evaluation set, these results suggest that a 0.4B-parameter encoder can approximate the performance of an ensemble of three 24–31B-parameter LLMs on claim identification at substantially lower inference cost.

\begin{table}[t]
  \centering
  \footnotesize
  \setlength{\tabcolsep}{3pt}
  \begin{tabular}{l ccc c}
    \toprule
    Model & P & R & F1 & $\kappa$ [95\% CI] \\
    \midrule
    Qwen3.6-27B
      & \textbf{.962} & .673 & .792 & .70 [.61, .78] \\
    Gemma-4-31B
      & .771 & \textbf{.956} & .854 & .75 [.67, .82] \\
    Mistral-3.2-24B
      & .832 & .832 & .832 & .73 [.64, .81] \\
    maj3
      & .888 & .841 & .864 & .78 [.71, .85] \\
    \midrule
    SciBERT$^\dagger$
      & .926 & .811 & .865 & .79 [.71, .85] \\
    DeBERTa-v3-large$^\dagger$
      & .939 & .823 & \textbf{.877}
      & \textbf{.81 [.75, .88]} \\
    \bottomrule
  \end{tabular}
  \caption{Claim identification on the 294 \textit{gold} sentences. F1 score is binary F1 for the claim class, $\kappa$ is Cohen's $\kappa$ against the gold label. \textit{maj3} denotes the majority label of the three LLMs. $^\dagger$Fine-tuned only on \textit{maj3} silver labels.}
  \label{tab:llm}
\end{table}

\subsection{Error Analysis} 
\label{ss:error_analysis}
All three LLMs jointly missed only 2 of the 113 gold-standard claims and jointly misclassified only 2 of the 181 non-claims, allowing majority voting to mitigate model-specific errors.
%
Qwen, the most conservative model (\autoref{fig:core_eval}), showed substantially lower recall for claims in abstracts and conclusions (0.42) than for those in results and subsequent sections (0.76). Relying on Qwen alone could therefore lead to underrepresentation of authors' main findings.
%
In contrast, DeBERTa shows high agreement with its teacher ensemble, \textit{maj3}, matching its predictions on 272 of the 294 gold-standard sentences.
%
Of the 30 sentences misclassified by \textit{maj3}, DeBERTa corrects 13, primarily procedural or factual statements incorrectly classified as claims by Gemma and Mistral, while introducing only 9 new errors among the 264 sentences correctly classified by \textit{maj3}.
%
On the 60 \textit{core} sentences, DeBERTa achieves perfect agreement with the annotator majority ($\kappa = 1.00$), compared with $\kappa = 0.81$--$1.00$ (mean: 0.89) for individual annotators against the majority of their peers.
%
Finally, the encoder appears to inherit its teacher's section-specific biases, achieving higher recall in results and subsequent sections (0.89) than in abstracts and conclusions (0.62).


\subsection{Out of Corpus Evaluation}

To assess cross-corpus generalization, we evaluate our DeBERTa model, fine-tuned exclusively on silver-standard data, on a preview version of ClaimFlow \citep{pramanick2026claimflow} without additional training.
%
The dataset contains 1,365 human-annotated central claims (1-5 per paper) and their corresponding evidence sentences from 388 ACL Anthology papers.
While ClaimFlow does not require claims to be entity-anchored, we observe substantial overlap: 91.7\% of its claims contain at least one task, dataset, or method entity in either their summary or evidence (rising to 98.6\% when generic mentions are included).
Furthermore, entities are highly indicative of evidence sentences; 77\% of ClaimFlow's evidence sentences contain an entity, compared to only 63\% of unselected sentences in the same sections. 
When evaluated zero-shot, our model flags 53.6\% of ClaimFlow's evidence sentences as claims. 
Because a single claim can be supported by multiple evidence sentences, the model successfully identifies 62.3\% of the central claims via at least one evidence sentence, a figure that increases to 70.4\% when evaluated alongside the claim summary text. 
%

\begin{table}[t]
  \centering
  \small
  \begin{tabular}{lrrr}
    \toprule
    \textbf{Set} & \textbf{Papers} & \textbf{Sentences} & \textbf{Claim Ratio} \\
    \midrule
    Core      & 32  & 60     & 38.3\% \\
    Gold            & 43  & 294    & 38.4\% \\
    Silver   & 50  & 9,073  & 16.4\% \\
    ClaimFlow & 388 & 14,418 & 14.8\% \\
    \bottomrule
  \end{tabular}
  \caption{Dataset statistics. \textit{Claim ratio} is the proportion of claim sentences. \textit{Gold} includes human labels, with \textit{Core} labeled by all annotators. \textit{Silver} uses 3-LLM majority voting (excluding the 300 gold sentences). \textit{ClaimFlow} preview dataset.}
  \label{tab:dataset_statistics}
\end{table}

\paragraph{Blind Validation} 
Our model identifies claims in 24.1\% of entity-bearing sentences and 14.0\% of sentences without entity mentions that ClaimFlow did not label as claim-bearing.
Because ClaimFlow is restricted to a few central claims per paper, measuring strict precision against its labels is not meaningful. 
%
To interpret the discrepancies, a human annotator performed a blind manual review of 100 sentences: 50 false positives, and 50 false negatives. 
%
Among the false positives, 78\% (95\% CI: 64\%-87\%) are genuine claims under our guidelines.\footnote{Sentences mixing a claim with a contribution statement were labeled as claims; excluding these reduces the rate to 73\%.} 
%
Conversely, of the false negatives, only 6\% (95\% CI: 2\%--17\%) qualify as claims under our schema. 
The vast majority of these misses are contribution statements (e.g., ``we propose \dots'', 20 of 47) or procedural descriptions (17 of 47), which our guidelines intentionally exclude. 
Therefore, the divergence between the datasets reflects a difference in definitions rather than model error.
ClaimFlow identifies a paper's central claims, whereas our model extracts an exhaustive set of entity-anchored, evaluative claims.

\section{Discussion}

\subsection{Does entity anchoring make claim annotation focused and reliable?} 
In general, every sentence in a scholarly paper can be considered as an assertion. 
For example, in Sci-Arg, full-text annotation produced an average of 205 claim spans per paper \cite{lauscher2018argument}.
%
However, entity anchoring reframes the annotation objective from a vague \textit{is this a claim?} to a specific \textit{does this sentence make a contestable assertion about this entity?}
This constraint significantly reduces the search space, as only 59\% of body sentences in our corpus contain scientific entities. 
It also maintains strictness, filtering out 62\% of these candidate sentences as non-claims to yield a manageable average of 32 claims per paper. 
More importantly, this localized focus enables highly reproducible annotations ($\alpha = 0.80$). 
Furthermore, the likelihood of a sentence constituting a claim increases with entity density (31\%, 37\%, and 51\% for one, two, and $\geq$3 entities, respectively), indicating that entities act as strong signals of argumentative content rather than arbitrary filters.
Although this approach inherently bounds the scope of extractable claims (see Limitations), it successfully transforms open-ended full-text annotation into a focused, reliable sentence-level decision.

\subsection{Can LLMs Replace Costly Full-Text Annotation?}
While entity anchoring reduces the annotation search space, manual full-text labeling remains impractical. 
Relying on a single LLM is also prone to errors due to varying model strictness and biases. 
%
However, as noted in the previous section, using an ensemble of models largely offsets these biases because their mistakes rarely coincide.
Consequently, our majority-vote ensemble (\textit{maj3}) achieved strong human agreement ($\kappa=0.78$ on \textit{gold}, $\kappa=0.82$ on \textit{core}).
We further validated this by training encoders on individual LLM outputs versus \textit{maj3}. 
%
Encoders trained on single-LLM outputs inherited their respective teacher's bias (Gemma: P 0.78 / R 0.95; Qwen: P 0.95 / R 0.71) and scored lower (F1 0.853 and 0.812) relative to the \textit{maj3}-trained DeBERTa (0.877). 
%
Furthermore, training encoders exclusively on our small human-annotated \textit{gold} set via 5-fold cross-validation produced weaker and less stable models (F1: 0.812 $\pm$ 0.019). 
Therefore, the LLM-generated \textit{silver} data provides a highly scalable alternative that yields a stable encoder performance.


\section{Conclusion}

We introduced \textbf{entity-anchored claim extraction}, an approach for extracting structured claims from full-text scholarly articles. 
By utilizing scientific entity mentions as anchors, our approach replaces fuzzy, document-level discourse modeling with well-defined, sentence-local claim extraction. 
We validated our methodology through a case study of 50 sentiment analysis papers published between 2013 and 2023. 
%
Our rigorous annotation pipeline yielded a reliable human \textit{gold} standard of 113 claim sentences, which we used to validate a large-scale \textit{silver} dataset of 1,487 claims generated via a majority-vote LLM ensemble. 
%
We demonstrated that encoders fine-tuned exclusively on our silver-standard dataset achieve more consistent claim identification performance than those trained on the smaller human-annotated gold-standard datasets or single-LLM-generated data.
%
These models also demonstrate strong zero-shot generalization to NLP papers outside the training corpus.
%

\section*{Limitations}

We acknowledge several limitations in our current study. 
First, our extraction pipeline relies exclusively on the GSAP-NER model, incorporating alternative or more exhaustive scientific entity recognition frameworks could potentially alter recall profiles. 
Second, our primary empirical validation is constrained to a densely linked case study of 50 sentiment analysis papers. 
Although our zero-shot experiments on the ClaimFlow dataset demonstrate generalizability across the broader NLP domain, the framework remains to be evaluated across more diverse scientific disciplines. 
Fundamentally, we recognize an inherent limitation of the entity-anchored paradigm: not every scientific argumentative component explicitly contains a typed scientific entity.
Consequently, our methodology does not guarantee an exhaustive extraction of all argumentative units within a publication. 
However, our primary objective is not exhaustive recall, but rather to drastically reduce the manual effort required to structure full-text arguments. 
Furthermore, our structured schema partially mitigates this restriction by capturing non-entity-bearing evidence within optional fields, such as the \textit{rationale}. 
Finally, the restriction to a 50-paper corpus was a deliberate choice to facilitate the dense, high-quality manual annotation of interlinked papers. 
Building on this foundation, our ongoing future work aims to model the complete argumentative graph, encompassing complex intra-document and cross-document relations.

\section{Ethical Considerations}
Our annotations are released under a CC BY-NC-SA 4.0 license, whereas quoted text fragments retain the original open access or publisher licenses of the respective venues (ACL Anthology, AAAI, ICWSM, IJCAI) for research use. Since transformer based models or LLMs can produce errors, automated claims should serve as assistive tools rather than definitive replacements for reading the original papers. LLM-generated silver labels may also inherit latent model biases. Compute requirements were minimal, limited to inference with three open-weight LLMs on a single GPU node and fine tuning lightweight encoders.

\paragraph{Use of AI:} We utilized AI assistants (Gemini and Claude) for code generation to develop our annotation tool, data analysis and to compute corpus statistics. All generated code and outputs were thoroughly reviewed, executed, and verified by the authors.

\section{Data and Code Release}
Our code, datasets, LLM prompts, and annotation guidelines are available on {\href{https://github.com/Bhuvanesh-Verma/entity-anchored-claim-extraction}{Github}}. A preview version of the ClaimFlow dataset was used with the authors' permission.

\bibliography{custom}

\appendix

\section{Corpus Construction and Preprocessing Details}
\label{app:corpus_stats}

\paragraph{Retrieval and Filtering.} The seed papers IMDb, SST, and SemEval-14 ABSA had 3,299, 6,781, and 1,886 forward citations in OpenAlex, respectively. We retrieved the most-cited citing papers for each (600, 400, and 600), yielding 1,508 unique papers after deduplication and the removal of review and survey articles. Keyword filtering (requiring \textit{sentiment, opinion, polarity, aspect,} or \textit{emotion} in the title or abstract) reduced this set to 876 articles. We downloaded the 397 available open-access PDFs, extracting up to the first 40 pages using \texttt{pdfplumber}, and retained only documents containing at least 1,500 words.

\begin{table}[hbt]
\centering
\small
\begin{tabular}{@{}lp{0.70\columnwidth}@{}}
\toprule
\textbf{Entity Type} & \textbf{Keywords} \\ \midrule
\textbf{Datasets} & 
IMDb, SST, SST-2, SST-5, Stanford Sentiment Treebank, SemEval-2014, SemEval-2015, SemEval-2016, Sentiment140, Yelp, Amazon Reviews, Twitter US Airline, MR dataset, Rotten Tomatoes, Laptop14, Restaurant14 \\ \addlinespace
\textbf{Methods} & 
SVM, Naïve Bayes, LSTM, BiLSTM, Bi-LSTM, GRU, CNN, RNN, BERT, RoBERTa, XLNet, DistilBERT, ALBERT, Transformer, Word2Vec, GloVe, FastText, TF-IDF, Attention, Memory Network, Capsule Network \\ 
\bottomrule
\end{tabular}
\caption{Curated keyword entities used for establishing cross-document entity edges.}
\label{tab:curated-keywords}
\end{table}

\begin{table}[t]
\centering
\small
\setlength{\tabcolsep}{4pt}
\begin{tabular}{lr}
\toprule
\textbf{Property} & \textbf{Value} \\
\midrule
Papers                          & 50 \\
Publication years               & 2013-2023 \\
Full-text length (median words) & 4,533 \\
Citations received (median)     & 191 \\
\midrule
\multicolumn{2}{l}{\textit{Papers citing each seed benchmark}} \\
\quad SemEval-14 ABSA           & 30 \\
\quad SST                       & 19 \\
\quad IMDb                      & 10 \\
\midrule
Paper pairs linked by citation  & 176 \\
Papers linked within the set    & 48 \\
\bottomrule
\end{tabular}
\caption{The 50 selected papers. All are open access with full text. A paper may cite more than one seed benchmark (nine cite two), so the seed counts sum to more than 50. Citation counts are from OpenAlex.}
\label{tab:selected}
\end{table}

\paragraph{Graph-Based Sampling.} In our graph construction, an entity edge was formed if two papers shared at least two keyword entities (from a curated list of 16 datasets and 21 methods shown in \autoref{tab:curated-keywords}), with at least one being a dataset. Using greedy modularity maximization (implemented via NetworkX), we selected the largest community, comprising 209 papers. To select the final 50 papers, we ranked them first by within-community citation edges, then by entity edges. We stratified the selection across four publication year bands:  2014 and before, 2015-2017, 2018-2020, and 2021+, targeting 12-13 papers per band. As the 2014 and before band contained only 8 papers, we backfilled by rank, resulting in 17 papers for 2015--2017 and 13 for 2018-2020. More information about the select 50 papers are shown in \autoref{tab:selected}.

\paragraph{Text Cleaning.} Following GROBID (v0.8.2) extraction, we applied localized text normalization. Specifically, we corrected hyphenation introduced by line breaks (e.g., ``inter- nal'' $\rightarrow$ ``internal'') only when bounded by lowercase letters, deliberately preserving acronym compounds (e.g., ``RAM- training''). We also repaired artifact hyphens stemming from PDF extraction errors (e.g., ``Contex-tAVG'' $\rightarrow$ ``ContextAVG'') and recurring tokenization splits (e.g., ``sof tmax'' $\rightarrow$ ``softmax''). Space formatting prior to punctuation was also standardized.


\section{Entity Extraction and Normalization Details}
\label{app:sci_entity}

\paragraph{Model Training and Performance.} 
We trained the SciDeBERTa-CS taggers using the original authors' hyperparameters. As a baseline validation, on fold~0 of the GSAP-NER dataset, our specific-type tagger achieved a micro-F1 score of 64.6 under exact match and 73.4 under partial match criteria.

\paragraph{Post-Processing and Span Resolution.}
Applied to our 50-paper corpus, the taggers initially produced 27,050 spans. We implemented a deterministic post-processing step to remove sub-word fragments, merge adjacent spans of the same entity type, and resolve overlapping duplicates. This reduction resulted in 23,832 valid mentions (detailed in Table \ref{tab:entity-labels}). 

\paragraph{Filtering and Normalization.}
From the resolved mentions, we retained 13,720 instances corresponding to the six specific entity types and 3,454 \textit{ReferenceLink} markers. To address surface-level variations of identical concepts (e.g., \textit{semeval 2014}, \textit{semeval-2014}, and \textit{semeval}), we applied a normalization heuristic. Following this normalization, the corpus encompasses 3,627 distinct scientific entities (see \autoref{tab:entity-types}).

\begin{table}[t]
\centering
\small
\setlength{\tabcolsep}{4pt}
\begin{tabular}{llr}
\toprule
\textbf{Label} & \textbf{Use} & \textbf{Mentions} \\
\midrule
Method            & entity          & 5,035 \\
ModelArchitecture & entity          & 2,849 \\
Task              & entity          & 2,656 \\
MLModel           & entity          & 2,080 \\
Dataset           & entity          &   980 \\
Datasource        & entity          &   120 \\
\midrule
ReferenceLink     & citation link   & 3,454 \\
MLModelGeneric    & discarded       & 3,313 \\
DatasetGeneric    & discarded       & 3,313 \\
URL               & discarded       &    32 \\
\midrule
\multicolumn{2}{l}{Specific / all mentions} & 13,720 / 23,832 \\
\bottomrule
\end{tabular}
\caption{GSAP-NER mentions in the 50 papers after post-processing.}
\label{tab:entity-labels}
\end{table}

\begin{table}[h]
\centering
\footnotesize
\setlength{\tabcolsep}{3pt}
\begin{tabular}{p{0.64\columnwidth}r}
\toprule
\textbf{Statistic} & \textbf{Value} \\
\midrule
Specific entity mentions & 13,720 \\
Per paper (median; range) & 250; 94-605 \\
Body sentences with $\geq$1 entity & 6,220 \\
Distinct entities (normalised) & 3,627 \\
\quad Method / Task & 2,027  / 488 \\
\quad MLModel / ModelArchitecture & 330 / 587 \\
\quad Dataset / Datasource & 172 / 23 \\
Entities in $\geq$2 / $\geq$5 / $\geq$25 papers & 674 / 161 / 11 \\
\bottomrule
\end{tabular}
\caption{Entity-layer statistics. The most widespread entities are
\textit{sentiment analysis} (40 papers), \textit{word embeddings} (38),
and \textit{ABSA} (33).}
\label{tab:entity-types}
\end{table}

\section{Annotation Interface and Schema Details}
\label{app:appendix_annotation_ui}

We developed a custom, lightweight web-based tool (Flask/SQLite) for blind annotation. 
To provide necessary context without overwhelming the reader, candidate sentences are presented in reading order embedded within a local context window (two preceding and two succeeding sentences), alongside the paper title and section heading. 
Distribution of stratified sampling of the sections for final annotation is shown in \autoref{tab:claim-sampling}.
To assist with structuring, automatically extracted scientific entities are highlighted in distinct colors by type, while generic entities (resolvable via the surrounding context) are underlined. 

The annotation process proceeds in two steps:

\begin{itemize}
    \item \textbf{Identification:} Annotators classify the target sentence as argumentative (\textit{claim}) or non-argumentative (\textit{non-claim}). A \textit{Cannot Judge} option is provided to gracefully handle and exclude garbled text resulting from PDF extraction errors or instances lacking sufficient context.
    \item \textbf{Structuring:} For identified claims, annotators specify the claim type (\textit{Own} or \textit{Reported}) and assign \textit{Subject} and \textit{Object} arguments by clicking the highlighted entities. If an appropriate entity was missed by the automated extraction, annotators are permitted to input free text. Finally, they select from a predefined inventory of 13 predicates (12 specific predicates plus an \textsc{other} category), which are fully detailed in \autoref{tab:predicates}.
\end{itemize}

\begin{table}[t]
\centering
\small
\setlength{\tabcolsep}{3pt}
\begin{tabular}{lrrrr}
\toprule
\textbf{Sections} & \textbf{Pool} & \textbf{Sampled} & \textbf{Rate} & \textbf{Core} \\
\midrule
Results onward$^\dagger$ & 1,293 & 175 & 13.5\% & 35 \\
Abs. + Conc.            &   381 &  75 & 19.7\% & 15 \\
Intro. + Rel. Work      & 1,101 &  35 &  3.2\% &  7 \\
Method + Exp.           & 1,396 &  15 &  1.1\% &  3 \\
\midrule
Total                   & 4,171 & 300 &  7.2\% & 60 \\
\bottomrule
\end{tabular}
\caption{Stratified sentence sampling for claim annotation across document sections. \emph{Core} denotes the overlapping subset used for inter-annotator agreement. Abbreviations: Abs. (Abstract), Conc. (Conclusion), Intro. (Introduction), Rel. Work (Related Work), Exp. (Experiment). $^\dagger$Includes Results and all subsequent pre-conclusion sections (e.g., discussion, error analysis).}
\label{tab:claim-sampling}
\end{table}

To accommodate complex sentences, the interface allows annotators to capture multiple independent claim records per sentence. 
Furthermore, optional fields including the claim's \textit{scope}, \textit{condition}, \textit{rationale}, \textit{attribution}, and specific evaluation metrics alongside their numeric results, can be explicitly extracted when stated in the text.

\begin{table*}[t]
\centering
\small
\setlength{\tabcolsep}{5pt}
\begin{tabular}{lp{8.2cm}l}
\toprule
\textbf{Predicate} & \textbf{Definition} & \textbf{Constraint} \\
\midrule
\texttt{outperforms}
& Subject achieves better task performance than the object on a shared task, dataset, or metric.
& object \\

\texttt{underperforms}
& Subject achieves worse task performance than the object on a shared task, dataset, or metric.
& object \\

\texttt{achieves}
& Subject attains a stated performance level, with no named comparison object.
& no object \\

\texttt{improves}
& Subject raises performance relative to a prior state, baseline, or component it modifies.
& object \\

\texttt{degrades}
& Subject lowers performance relative to a prior state, baseline, or component it modifies.
& object \\

\texttt{generalizes-to}
& Subject transfers to the object domain, task, language, or dataset while retaining performance.
& object \\

\texttt{fails-on}
& Subject performs unacceptably on the object input class or condition.
& object \\

\texttt{requires}
& Subject depends on the object (resource, annotation, component) to function or to reach its reported performance.
& object \\

\texttt{is-suited-for}
& Subject is appropriate or preferable for the object setting.
& object \\

\texttt{is-standard-for}
& Subject is the conventional or default choice for the object task or setting.
& object \\

\texttt{is-comparable-to}
& Subject performs equivalently to the object on a stated task, dataset, or metric; asserts no meaningful difference, not superiority.
& object \\

\texttt{is-more-efficient-than}
& Subject costs less than the object on a stated non-accuracy dimension; asserts nothing about output quality.
& object + dimension$^\dagger$ \\

\texttt{OTHER}
& The sentence is a claim but no predicate above fits; the annotator writes the predicate they would have used.
& free text \\
\bottomrule
\end{tabular}
\caption{Predicate inventory for describing a claim. \emph{Object}: the claim must name what the subject is compared against or directed at (an entity or free text). $^\dagger$\emph{Dimension}: one of speed, memory, parameters, data, compute, simplicity, or other.}
\label{tab:predicates}
\end{table*}




\section{Implementation Details}
\label{app:impl_details}
For experimental reproducibility, we detail our computational setup and hyperparameters. Entity extraction via the GSAP-NER tagger was executed on a single NVIDIA RTX 4060 Ti (16GB VRAM) utilizing CUDA 13, Python 3.13, PyTorch 2.13, and Transformers 5.14. For silver data generation via LLMs, we deployed Qwen3.6-27B utilizing \texttt{vLLM} 0.17.1, alongside Mistral-Small-3.2-24B and Gemma-4-31B-it utilizing \texttt{vLLM} 0.29.0, hosted on an NVIDIA H200 NVL (141GB) node. Inference utilized OpenAI-compatible chat completions with prompts formulated as a single user message, employing greedy decoding (temperature 0), a 2,048 maximum token limit, a 16,000 maximum context length (\texttt{max\_model\_len}), and with internal thinking modes strictly disabled. Finally, the encoder fine-tuning was performed on an NVIDIA L40S GPU. We run 5-fold cross-validation split by paper. The 43 papers that contain human-annotated sentences are divided into five folds balanced by annotated-sentence count (8-9 papers, nearly 60 sentences per fold). The remaining 7 papers are always used for training. In each fold, the model is trained on silver labels (majority vote of three LLMs) from the other papers, with 10\% of them (4 papers) held out as a development set for early stopping. It is evaluated on the human labels of the held-out fold. No paper appears in more than one of train, dev and test, and no human-annotated sentence is used in training, even with a silver label. We optimized the models using AdamW for a maximum of 5 epochs with a batch size of 32, a maximum sequence length of 128, a learning rate of $1 \times 10^{-5}$, a weight decay of 0.01, a 10\% linear warmup schedule, and a gradient clipping norm of 1.0. Training was regulated via early stopping based on the development set positive-class F1 score, utilizing a patience of 2 epochs and restoring the best-performing weights.



\end{document}